%% file: Manipulation_GAN.tex
\documentclass[journal]{IEEEtran}
\usepackage{graphicx}
\usepackage{amsmath}
\usepackage{verbatim}
\usepackage{amssymb}
\usepackage{float}
\usepackage{comment}
\usepackage{caption}
\usepackage{hhline}
\usepackage{rotating}
\usepackage{multirow}
\usepackage{longtable}

\usepackage{color}

\DeclareMathOperator{\E}{\mathbb{E}}

\usepackage{cite}

\ifCLASSINFOpdf
\else
\fi
\usepackage{amsmath}
\usepackage{url}

\input{Formatting/notation.tex}

\begin{document}
%
\title{A Transferable Anti-Forensic Attack on Forensic CNNs Using A Generative Adversarial Network}
%
%

\author{Xinwei~Zhao,~\IEEEmembership{Student Member,~IEEE},  Chen~Chen,
        Matthew~C.~Stamm,~\IEEEmembership{Member,~IEEE}
        
\thanks{This material is based upon work supported by the National Science Foundation under Grant No. 1553610. Any opinions, findings, and conclusions or recommendations expressed in this material are those of the authors and do not necessarily reflect the views of the National Science Foundation.}
\thanks{This material is based on research sponsored by DARPA and Air Force Research Laboratory (AFRL) under agreement number PGSC-SC-111346-03. The U.S. Government is authorized to reproduce and distribute reprints for Governmental purposes notwithstanding any copyright notation thereon. The views and conclusions contained herein are those of the authors and should not be interpreted as necessarily representing the official policies or endorsements, either expressed or implied, of DARPA and Air Force Research Laboratory (AFRL) or the U.S. Government.}
\thanks{The authors are with the Department of Electrical and Computer Engineering, Drexel University, Philadelphia, PA, 19104 USA (e-mail: xz355@drexel.edu; cc93ustc@gmail.com; mstamm@drexel.edu).
}}

%
%

\markboth{Journal of \LaTeX\ Class Files,~Vol.~X, No.~Y,  MONTH~YEAR}%
{Shell \MakeLowercase{\textit{et al.}}: Bare Demo of IEEEtran.cls for IEEE Journals}
%



\maketitle

\begin{abstract}
\input{sections/abstract_v2_MCS.tex}
\end{abstract}


%
\IEEEpeerreviewmaketitle
\vspace{-1em}
\section{Introduction}
\input{sections/intro_v2_MCS.tex}

\section{Image Forensic CNNs}
\input{sections/imageCNNs.tex}

\section{Problem Formulation}
\input{sections/problem_2_MCS.tex}

\section{Knowledge Scenarios}
\input{sections/scenarios}

\section{Proposed Anti-Forensic Attack}
\input{sections/proposed.tex}

\section{Evaluation Metrics}
\input{sections/eval_metrics_v2_MCS.tex}

\label{sec: experiments}
\section{Experimental Results}
\input{sections/experiment_setup_v2_MCS.tex}

\input{sections/perfect.tex}

\input{sections/data_mismatch.tex}

\input{sections/para_mismatch_v2_MCS.tex}

\input{sections/arch_mismatch_v2_MCS.tex}

\section{Conclusion}
\input{sections/conclusion.tex}

\ifCLASSOPTIONcaptionsoff
  \newpage
\fi




\bibliographystyle{IEEEtran}
\bibliography{refs/citations}

\end{document}

%% file: Formatting/notation.tex
\newcommand*{\rom}[1]{\uppercase\expandafter{\romannumeral #1\relax}}

\newcommand{\subhead}[1]{ 
	\vspace{0.2em}
	\noindent \textbf{{\smash{#1}}}:  }

%% file: sections/abstract_v2_MCS.tex
With the development of deep learning,  convolutional neural  networks (CNNs) have become widely used in multimedia forensics for tasks   such as  detecting and identifying image forgeries.   
Meanwhile,  anti-forensic attacks have been developed to fool these CNN-based forensic algorithms.  
Previous anti-forensic attacks often were designed to remove 
forgery traces left by a single manipulation operation as opposed to a set of manipulations.  
Additionally,  recent research has shown that existing anti-forensic attacks against forensic CNNs  
have poor transferability, i.e. they are unable to fool other forensic CNNs that were not explicitly used during training.
In this paper,  we propose  a new  anti-forensic attack framework designed to  remove  forensic traces left by a variety of manipulation operations.  
This attack is transferable, i.e. it can be used to attack forensic CNNs are unknown to the attacker, and it introduces only minimal  distortions that are imperceptible to human eyes.
%
Our proposed attack utilizes a generative adversarial network (GAN) to build a generator that can attack color images of any size.  
We achieve attack transferability through the use of a new training strategy and loss function.  
We conduct extensive 
 experiment to demonstrate 
that our attack can fool many state-of-art forensic CNNs with varying levels of knowledge available to the attacker. 

%% file: sections/intro_v2_MCS.tex
Software, such as PhotoShop and GIMP, makes photo editing easy for people who have no background in image processing and allow people to add or remove contents and effects of preference. In many critical scenarios, however, multimedia contents  can be used as digital evidence to assist in activities such as criminal investigations and news reporting. Therefore, it is important for forensic investigators to ensure  the integrity and authenticity of digital contents~\cite{stamm2013information, piva2013overview, gloe2007can,milani2012_vidForensOverview}. To combat  multimedia forgeries, researchers have  developed various forensic algorithms  to detect image forgeries~\cite{fridrich2003detection, chen2008determining,  bayram2006image, bayram2009efficient, JPEG_ghosts, amerini2011sift, kee2011digital,  pan2010region,  bianchi2012image, mayer2018accurate},  and identify manipulation operations,  such as resizing~\cite{kirchner2008fast, popescu2005exposing}, contrast enhancement~\cite{stamm2010forensic, piva2013overview}, JPEG compression~\cite{ huang2010detecting, bianchi2011detection, barni2016adversary}, median filtering~\cite{yuan2011blind} and etc. In recent years,  deep-learning based techniques  such as CNNs have become the most popular approach for sophisticated and robust forensic algorithms and achieved many  most state-of-art performances in digital forensics~\cite{mayer2020similarity, MISLNet, bayar2018towards, barni2018cnn, splicebuster, bondi2017tampering,  kang2013robust, noiseprint, bondi2016first,Tuama2016CNN}.  

To help investigators discover weaknesses of forensic techniques and develop defense strategies,  it is equivalently important to study anti-forensics~\cite{stamm2012forensics, barni2018adversarial}.  In some scenarios,  an intelligent attacker may launch a malicious attack to fool forensic algorithms by removing the forensic traces left by manipulations~\cite{sharma2016wifsanti, fontani2012hiding, kirchner2008_hiding_resampling, antiContrast2010, stamm_dithering}.  Previous research  has shown that the deep learning based algorithms  are vulnerable to adversarial examples~\cite{nguyen2015deep, DeepFool_2016_CVPR, 2014arXiv1412.6572G,  szegedy2014intriguing, carlini2017adversarial, papernot2016limitations, chen2018ead, su2019one, eykholt2018robust, brown2017adversarial, hayes2018visible, athalye2017synthesizing, kurakin2016adversarial}.  Recent research on digital anti-forensics has shown that attacks can be crafted to fool various forensic CNNs~\cite{kim2018medianGAN, Mislgan, chen2019generative, yu2016multi, counter2017CNN}.  

While these attacks can achieve strong 
performance when they are explicitly trained against a victim CNN (i.e white-box scenarios),  forensic researchers have found  that many anti-forensic attacks cannot fool other CNNs other than those used directly during training~\cite{transferability_Barni, zhao2020effect}.  This common problem is addressed as transferability issues of adversarial attacks.  The transferability issue of  anti-forensic attacks occurs in limited knowledge scenarios when the attacker has no direct access to the trained victim CNNs (i.e. the CNNs that they wish to attack).  The transferability of attacks has been actively studied in machine learning fields,  such as computer vision~\cite{liu2016delving, papernot2016transferability, carlini2017adversarial}.  However,  very limited research has been done to address this problem in forensics.  Recent research has  demonstrated that existing anti-forensic  attacks on CNNs,   such as FGSM and GAN-based attack,  have many difficulties in transferring~\cite{transferability_Barni, zhao2020effect}.  Particularity,  Barni et. al  has shown that attacks on image forensic CNNs have difficulty transferring to other CNNs constructed with different architectures or trained using different training data~\cite{transferability_Barni}.  Moreover,  Zhao et. al found that by simply changing the definitions of classes of a image forensic CNN,  the adversarial examples cannot fool anecdotally same CNNs~\cite{zhao2020effect}. 

In this paper, we propose a new anti-forensic attack based on generative adversarial networks (GANs) to fool forensic CNNs trained for manipulation detection and identification.  
Our proposed attack operates by using a generative model to remove forensic traces left by  manipulation operations from an image,  and also synthetically constructs forensic traces of unaltered images.  The anti-forensically attacked image produced by our proposed attack can mimic the forensic statistics of real  unaltered  images and fool state-of-art forensic CNNs under different scenarios.  
Moreover,  our proposed attack demonstrates strong transferability under various limited knowledge or black box scenarios.  This includes transferring across different training data and CNN architectures.  
%
Our proposed anti-forensic GAN differs from the traditional GAN structure by incorporating additional elements.  
%
To achieve transferability, our 
proposed anti-forensic GAN introduces an ensemble of pre-trained surrogate CNNs  into training the generator.  This  forces the generator to learn more generic forensic traces of unaltered images.  
Each surrogate CNN is trained by the attacker to perform manipulation detection or identification,  and forms an unique decision boundary of the unaltered class.  By using an ensemble of surrogates CNNs constructed with diverse architectures and class definitions,  the generator is trained to capture a comprehensive  forensic information of unaltered images,  and produces anti-forensically attacked images residing on the intersection of boundaries of unaltered class formed by each surrogate CNN.  Additionally,  we introduce the rule of pixel to pixel correspondence for creating training data  to improve the transferability of the proposed attack. 

We evaluate our proposed anti-forensic GAN attack through  extensive amount of experiments  different knowledge levels of the investigator's CNN,  including perfect knowledge scenarios and three limited knowledge scenarios.  We demonstrate that our proposed anti-forensic GAN attack can fool forensic CNNs  built with various state-of-art architectures.  Moreover,  our proposed anti-forensic GAN attack  shows  strong transferability,  and can fool CNNs built with different architectures and trained on different database.  Additionally,  our proposed attack  will not introduce  visual  distortions  to the produced anti-forensically attacked images under any scenario.

%% file: sections/imageCNNs.tex
\label{sec: CNNs}
Convolutional neural networks (CNNs) are one of most popular deep learning frameworks.  
They have achieved  state-of-art performances on many forensic tasks,  including detecting and identifying manipulation operations~\cite{BayarIHMMSec2016_ManpiDetCNN,bayar2018towards,  bondi2017cluster, barni2018cnn}.  Image forensic CNNs operates by 
learning forensic feature extracted from a large amount of training data,  then forming decision boundary for each pre-defined forensic class.    


There are multiple ways to define classes of image forensic CNNs.
Zhao et al.  specify three major class definitions in previous research:  an binary decisions of unaltered or manipulated,  multi-class definitions of unaltered vs several individual manipulations,  or multiclass definitions of unaltered vs. several parameterized versions of individual manipulations~\cite{zhao2020effect}.   
We now briefly discuss these class definitions.

\underline{Manipulation detection} is a binary classification between manipulated and unaltered.  One class is assigned to unaltered images,  and the other class is assigned to images manipulated in any form.  This class definition allows investigator to detect if manipulation was applied.

\underline{Manipulation classification} is a multi-class classification to identify a specific manipulation or unaltered.  One class is assigned to unaltered images,  and other classes are each assigned to individual manipulation operation.  For each individual  manipulated class,  all parameters related to this manipulation operation are grouped together into a single class. This class definition not only allows the investigator to detect if manipulations exist,  but also to identify which specific manipulation was applied. 

\underline{Manipulation parameterization} is a multi-class classification to identify a specific  manipulation and parameter pairing or unaltered.   One class is assigned to unaltered images,  and other classes are each assigned to each pairing of manipulation and parameter (or range of parameterizations).  This class definition could be used if the investigator wants to know very detailed information about a possible forger or identify inconsistencies in editing within an image.

Each of the above class definitions includes the \textit{``unaltered"} class.  To fool forensic CNNs,  the anti-forensically attacked image should be produced within the decision boundary of ``unaltered" class.   However,  the decision boundaries of CNNs  highly depend on the training data,  CNN architecture,   and the definition of classes.  Any change may result in the change of the decision boundaries.  Therefore,  many existing anti-forensic attacks against CNNs,  such as FGSM~\cite{szegedy2014intriguing} and MISLGAN~\cite{Mislgan},  can only fool CNNs directly trained against~\cite{transferability_Barni, zhao2020effect}.

%% file: sections/problem_2_MCS.tex
\label{sec: problems}




To formulate the anti-forensic attack proposed in this work, 
we begin by examining the interaction between an information attacker Alice and a forensic investigator Bob.  In this interaction, Bob will be confronted with an image and will attempt to determine if the image is unaltered or if it has been manipulated.  To do this,  he will make use of a CNN classifier $C(\cdot)$ that is trained to differentiate between unaltered images and manipulated ones.  Particularly,  we assume he will use manipulation detection,  manipulation identification or manipulation parameterization to define the classes,  and a class corresponding to \textit{``unaltered"} will always be present.  

The attacker Alice will use a photo editor to create a falsified image by applying image manipulations $M(\cdot)$ to an unaltered image $I$.  
Since Alice is aware that Bob will use his forensic CNN $C$ to determine if Alice's image is unaltered or not,  Alice will use an anti-forensic attack $A(\cdot)$ to remove forensic traces left by her manipulations.  Alice's primary goal in designing this attack is  to fool Bob's forensic classifier such that
\begin{equation}
C(A(M(I)))= \mbox{\textit{``unaltered"}}.
\end{equation}
Additionally, Alice wants to ensure that her anti-forensic attack does not introduce visually detectable distortions into her image, thus rendering it visually implausible.

For the purposes of this work,  we assume Alice can collect a large amount data and train deep learning based algorithms.   We assume that Alice's manipulations $M$ consist of a single editing operation,  however this work can be extended with minor modifications to consider multiple editing operations.  Furthermore, we assume that Alice will choose between $N$ potential manipulations,  all of which are known to Bob.  While in reality Bob may not know all possible manipulations available to Alice, this assumption corresponds to a worst-case-scenario for Alice and ensures that our attack can still be successful in unfavorable conditions.

%% file: sections/scenarios.tex
\label{sec: scenarios}
\begin{table*}[t!]
	\centering
	\caption{Match and mismatch information between the attacker and the investigator under each scenario. }
	\begin{tabular}{|l|cccc|}
		\hline
		\textbf{Knowledge Scenarios}& \textbf{Training Data} & \textbf{Manipulations } &\textbf{Parameters}& \textbf{Trained CNN}\\\hline
		Perfect knowledge & Match & Match& Match & Match\\
		Training Data Mismatch & Mismatch & Match & Match &Mismatch\\
		Training Data and Manipulation Parameter Mismatch & Mismatch& Match & Mismatch&Mismatch\\
		Training Data and CNN Architecture Mismatch & Mismatch & Match& Match&Mismatch\\\hline
	\end{tabular}
	\label{tab: scenarios}
\end{table*}

CNNs possessed by the investigator  Bob are the ``victim CNNs"  that the attacker Alice attempts to fool.  It is reasonable to assume that  Alice will use as much  information as possible to make the attack successful.   Based on the amount of  knowledge available to Alice,  it is common in multimedia anti-forensics to categorize  attack scenarios  into the perfect knowledge scenario and  limited knowledge scenarios~\cite{barni2018adversarial, transferability_Barni, zhao2020effect}.  In the perfect knowledge scenario,  we assume Alice has the same amount of  knowledge that Bob has (i.e whatever Bob has access to,  Alice has the equal access).  All other scenarios are considered as limited knowledge scenarios.   Since the conditions for limited knowledge scenarios are harder to formulate than the perfect knowledge scenario,   we formulate three limited knowledge scenarios as below,  and discuss Alice's  known and unknown information in each scenario.   

\subhead{Perfect Knowledge}
We assume Alice  have full access to the victim CNN trained and possessed by Bob,  or she is capable of reproducing an identical copy of the victim CNN to train her attack,  such as when Alice has access to Bob's training data and knows the details of Bob's CNN architecture. 

\subhead{Training Data Mismatch}
We assume Alice has no  access to the victim CNN trained and possessed by Bob,  nor she has access to Bob's training data.  However,  we assume Alice knows about architecture of the victim CNN.  She also knows about the manipulations and chosen parameters that Bob used to create his training data.    As a result,  Alice can train a ``well-approximated copy" of the victim CNN by using the same CNN architecture and her own collected data.  However, due to the discrepancy between the two sets of training data,  boundaries  formed by Bob's  CNN and Alice's CNN are not identical.  Therefore,  Alice's attack should transfer across training data to fool the victim CNN.

\subhead{Training Data and Manipulation Parameter Mismatch}
We assume Alice has no  access to the victim CNN trained and  possessed by Bob,  nor she has access to Bob's training data.  However,  we assume Alice knows about the architecture of the victim CNN.  She also knows about the manipulations of Bob's interest and a subset of parameters  of these manipulations.   As a result,  Alice can obtain a ``bad-approximated copy" of the victim CNN by using the same CNN architecture and her own collected data.  However,  due to the incomplete information about manipulation parameters,  the discrepancies between Bob's  CNN and Alice's CNN are bigger than the ``Training Data Mismatch" scenario stated above.  Therefore,  it is more challenging for Alice to  make her attack  transfer to  fool Bob's CNN,  especially to fool Bob's manipulation parameterization classifier,  since Bob's manipulation paramerization classifier can output more classes than Alice. 

\subhead{Training Data and CNN Architecture Mismatch}
We assume Bob uses a CNN constructed with a private architecture,  and Alice  cannot gather any information about Bob's CNN architecture by all means.  However,  we assume Alice  knows about the manipulations and chosen parameters that Bob used to create his training data.    Therefore,  Alice's trained attack should be able to transfer across both training data and CNN architectures to fool Bob's CNN. 

We summarize the match and mismatch information between the attacker Alice and the investigator Bob  for each knowledge scenario stated above in Table~\ref{tab: scenarios}.

%% file: sections/proposed.tex
\label{sec: proposed}
Our goal is to create anti-forensically falsified images to fool manipulation forensic CNNs,  such that the victim CNN will classify forged images produced by our attack as unaltered or authentic.  Additionally,  the anti-forensically attacked image produced by our attack should  contain no perceptible distortion.  To do it, we propose a new generative attack to remove forensic forgery traces left by manipulation operations from an image  and reconstruct falsified forensic traces of unaltered images.  Our proposed attack utilizes a generative adversarial network to build a generator that  will take in a forgery image of any size as input,  and produce an anti-forensically attacked image of the same size and same contents as the forgery image. 

\vspace{-1em}
\subsection{Generative Adversarial Networks}
Generative adversarial networks (GANs) are a deep learning framework and often used to produce visually realistic images in computer vision~\cite{GAN2014}.  A  GAN typically consists of two networks,   a discriminator $D(\cdot)$ and a generator $G(\cdot)$.  While the goal of the discriminator  is to distinguish between generated data and real data,  the goal of the generator  is to produce fake data that can mimic statistical distribution of real data  to fool the discriminator.  The mechanism of a GAN is to train the discriminator and the generator in an alternative fashion using  a minmax function~\eqref{eq: gan},  until the two networks reach an equilibrium. 

Assuming real images $I$ have distribution $I\sim p_r(I)$ and generated images $I'$ have distribution $I'\sim p_g(I')$,  the minmax function for training GAN  is formulated as, 
\vspace{-0.2em}
\begin{equation}
	\min_{G}\max_{D} \E_{I\sim p_r(I)}[\log D(I)]+\E_{I'\sim p_g(I')}[\log(1- D(I'))]
	\label{eq: gan}
\end{equation}
where $\E$ represents the operation of calculating expected value. 

\begin{figure}[t]
	\centering
	\includegraphics[scale=0.4]{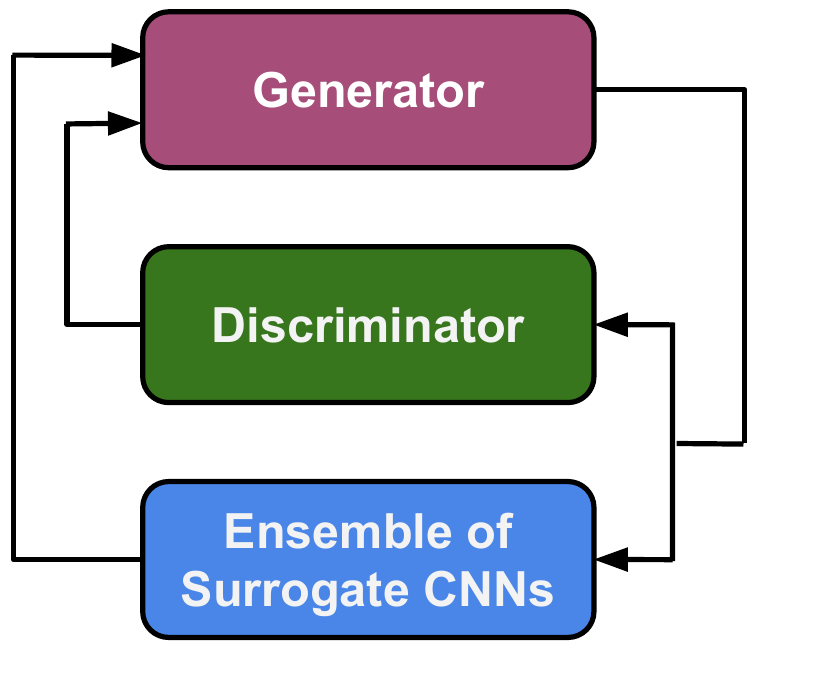}
	\caption{Overview of the proposed anti-forensic GAN framework.}
	\label{fig: anti-forensic_GAN}
\end{figure}

\vspace{-1em}
\subsection{Proposed Anti-Forensic GAN}
While GANs have been widely used to construct spatial contents and achieved many state-of-art performances~\cite{brock2018large, karras2019stylebased, zhu2020unpaired, isola2018imagetoimage},  limited research has been done to construct forensic traces to fool CNN-based forensic algorithms.  Chen et al.  proposed MISLGANs to fool forensic CNNs for camera model identification~\cite{Mislgan, chen2019generative}.   Kim et al.   showed that GANs can be used to remove forensic traces left by median filtering~\cite{kim2018medianGAN}.   However,  these algorithms were designed to remove particular forensic traces,  and there exists no anti-forensic attack that can remove forgery traces left by varying manipulation operations.  
Additionally,  previous research has shown that existing anti-forensic attacks on forensic CNNs,  such as FGSM~\cite{szegedy2014intriguing} and GAN-based attacks~\cite{Mislgan},  can only successfully fool those CNNs directly used to train the attack~\cite{transferability_Barni, zhao2020effect}.  This is a well-known problem of adversarial attacks referred as the transferability issue in machine learning fields.  While the transferability of attacks have been actively studied in computer vision~\cite{papernot2016transferability,  liu2016delving,Papernot_2017},  related research in multimedia anti-forensics is still in early stage.  

In this paper,  we propose a new GAN-based anti-forensic attack,  ``anti-forensic GAN'',  to solve the above problems.  The proposed anti-forensic GAN attack is a general anti-forensic  framework aimed to remove forgery traces left by various manipulation operations.  Additionally,  our proposed anti-forensic GAN attack are designed to transfer and can  fool CNNs not used during training.

Figure~\ref{fig: anti-forensic_GAN} shows the overall framework of the proposed anti-forensic GAN. 
Different from traditional GAN,  the proposed anti-forensic GAN consists of three major components,  a generator,  a discriminator  and an ensemble of surrogate CNNs.    The ensemble of  surrogate CNNs contains a diverse set of pre-trained forensic CNNs constructed with various CNN architecture and class definitions.  Each surrogate CNN  is trained for manipulation detection,  manipulation identification or manipulation parameterization.  The goal of the ensemble of surrogate CNNs is to  guide the generator to produce robust anti-forensically falsified images that can mimic comprehensive aspects of forensic information of unaltered images. 

We now describe the architecture of each component in the proposed anti-forensic GAN. 

\begin{figure}[t]
	\centering
	\includegraphics[scale=0.4]{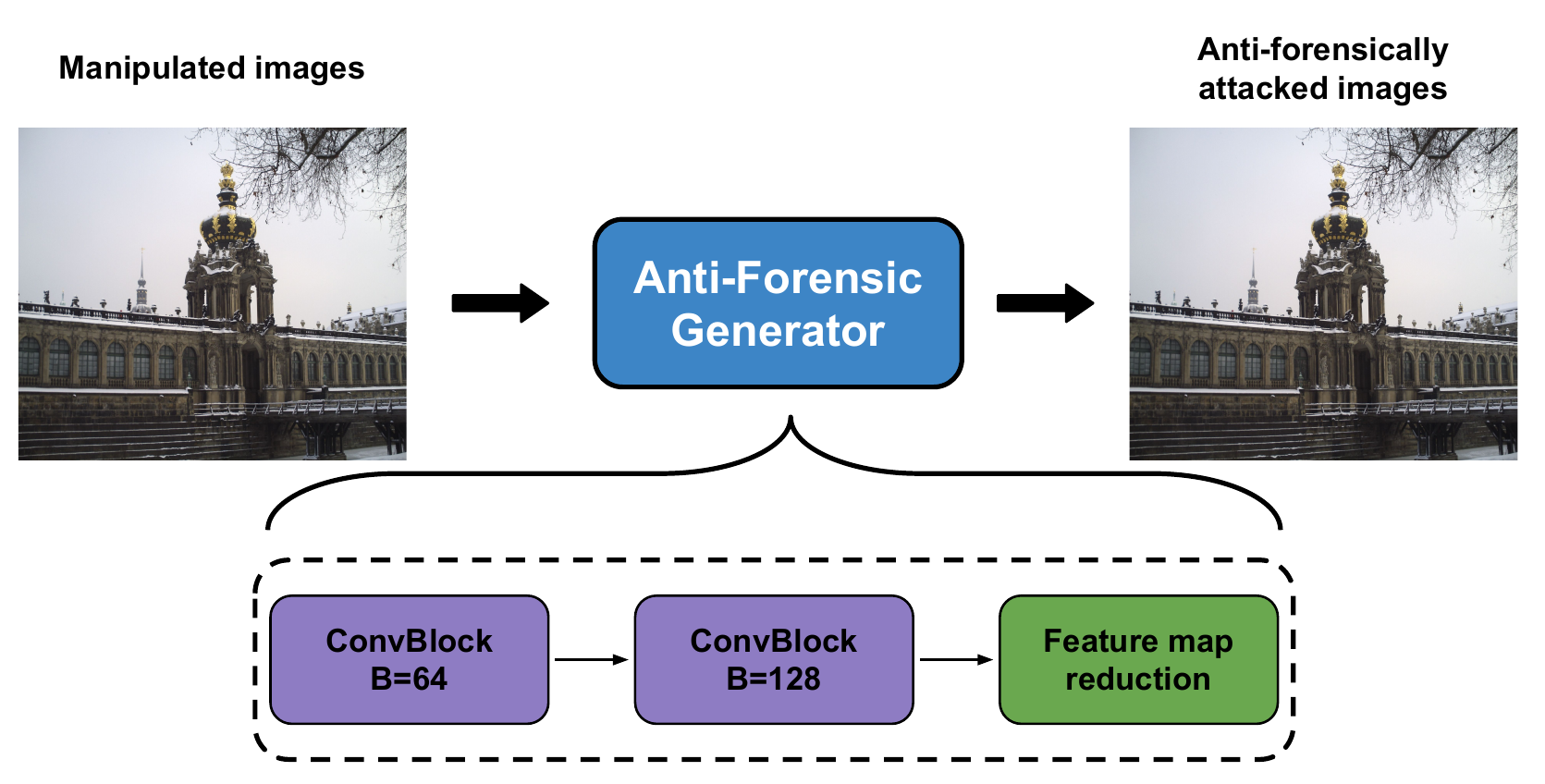}
	\caption{ The generator's architecture  of the proposed anti-forensic GAN framework.}
	\label{fig: generator}
\end{figure}

\begin{figure}[t]
	\centering
	\includegraphics[scale=0.4]{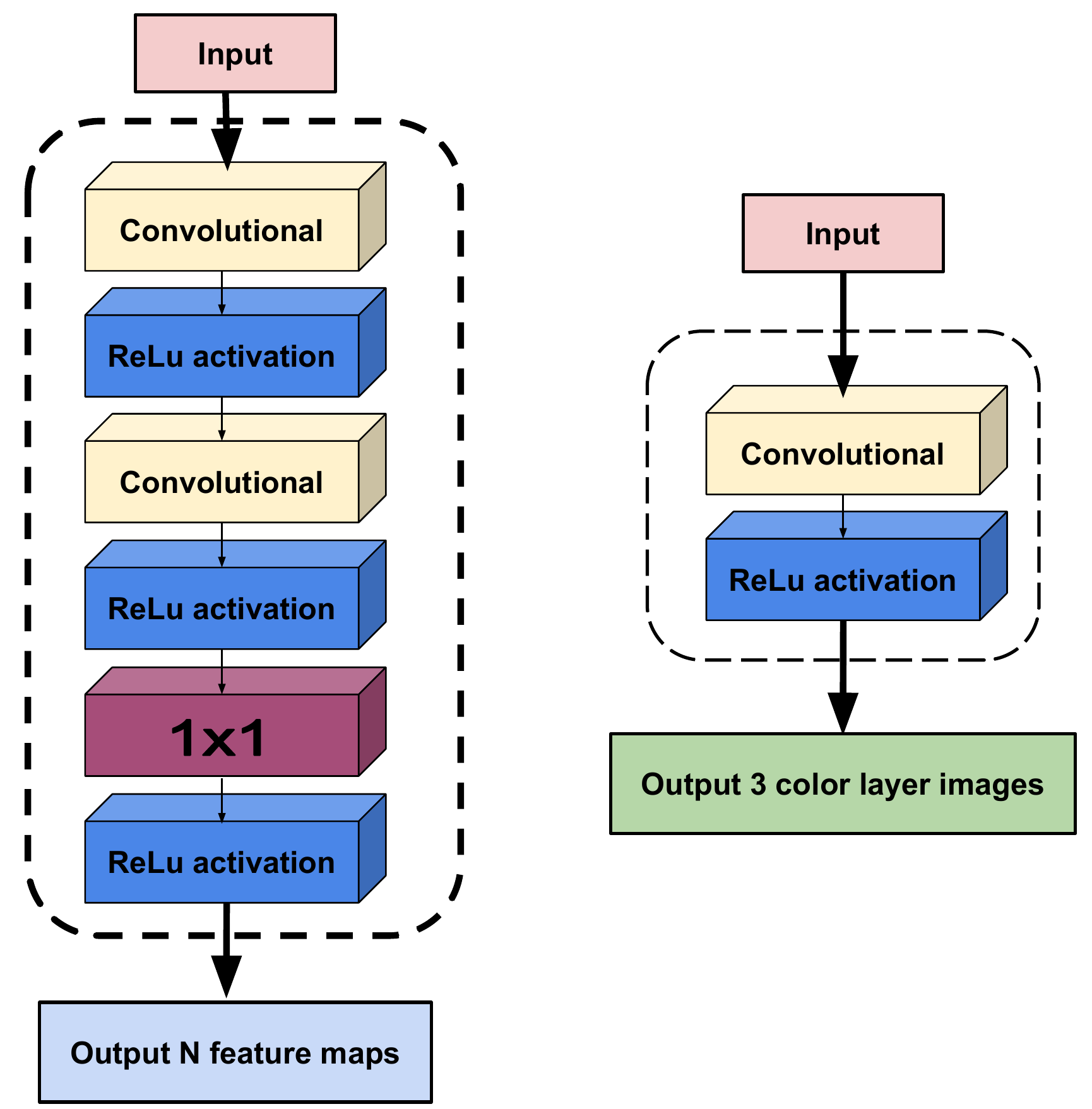}
	\caption{Building blocks of the generator:  convblock is shown on the left,  and feature map reduction is shown on the right.}
	\label{fig: basic}
\end{figure}

\subhead{Generator}
The goal of the generator is to remove forensic  traces left by manipulations and reconstruct the forensic traces to mimic forensic traces of unaltered images.   Figure~\ref{fig: generator} shows the architecture of the generator in the proposed anti-forensic GAN framework.  The generator consists of two major building blocks,  ConvBlocks and feature map reduction module,  shown in  Figure~\ref{fig: basic}. The ConvBlocks are conceptual blocks which are constructed with a sequence of convolutional layers with activation formed in common structure.  A $B=b$ ConvBlock is constructed with one convolutional layer with $b$,  $3 \times 3$ filters,  stride $1$  and ReLU activation~\cite{ReLU},  followed by another convolutional layer of the same structure,  and followed by a $1\times 1$ convolutional layer with ReLU activation.  The purpose of using $1 \times 1$ convolutional layers is to enforce the generator to learn cross-channel correlations between feature maps.    Our generator used  two consecutive ConvBlocks,  $B=64$  for the first one and $B=128$ for the second one.  

Since the second ConvBlock will output a large amount of feature maps,  we use the feature map reduction module to combine all the feature maps into a three-layer color image.  The feature map reduction module is constructed with a convolutional layer with three,  $3\times 3$ filters and stride $1$ followed by ReLU activation.

\begin{figure}[t]
	\centering
	\includegraphics[scale=0.35]{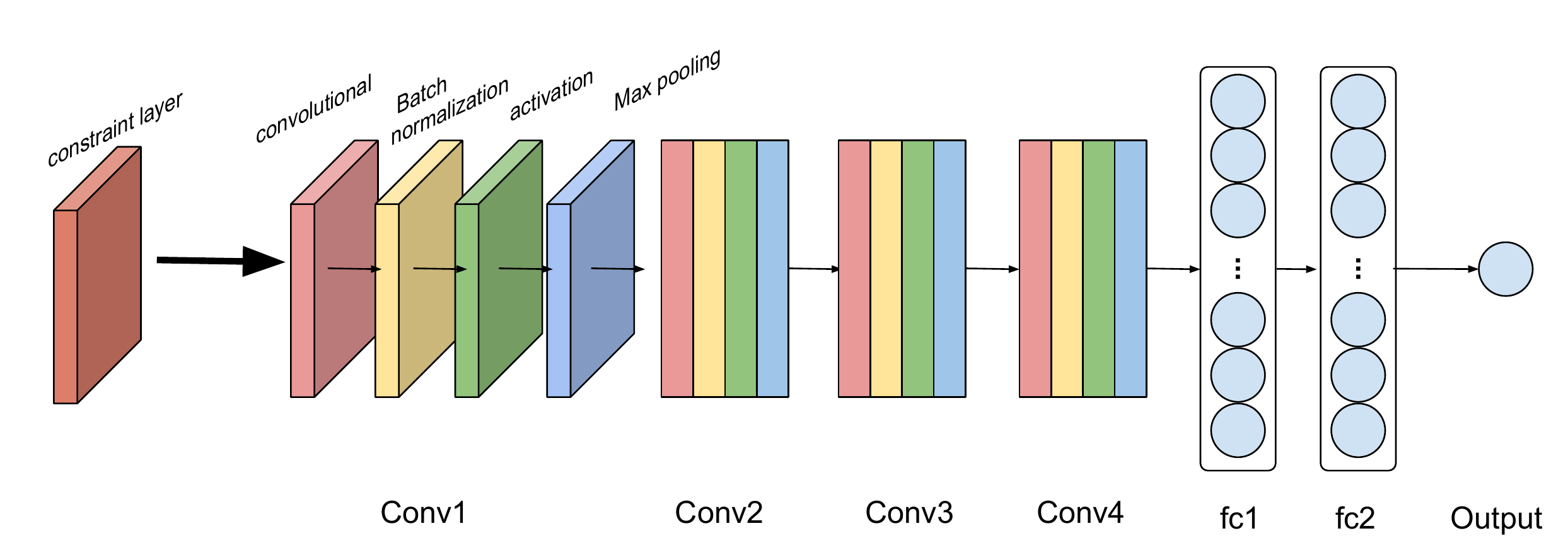}
	\caption{The discriminator's architecture of the proposed anti-forensic GAN framework.}
	\label{fig: discriminator}
\end{figure}

\subhead{Discriminator} 
The goal is the discriminator is  to differentiate the  unaltered images from  the generated anti-forensically attacked images.   The architecture of the discriminator is shown in Figure~\ref{fig: discriminator}.  It is  a variant of the  image forensic CNN proposed by Bayar and Stamm~\cite{bayar2017design},  since this CNN was designed to learn  forensic features  rather than the images' contents.  It first uses  a constrained convolutional layer to learn low level forensic features,   then uses four standard convolutional layers with batch normalization~\cite{batch_normalization} and hyperbolic tangent activation followed by three fully connected layers to extract high-level features.   We modify the last fully connected  layer to  a single neuron followed by sigmoid activation to construct our discriminator.  The last neuron activation corresponds to the likelihood that an image is unaltered or generated  (i.e 1 if the image is real,  0 if the image is generated).   

\subhead{Ensemble of Surrogate CNNs}
The goal of the ensemble of surrogate CNNs is to enforce the generator to produce anti-forensically attacked images that can mimic the forensic traces of unaltered images.   Each surrogate CNN is pre-trained to perform  manipulation detection,  identification or parameterization and can form an unique decision boundary of the unaltered class.  By integrating  diverse surrogates CNNs,  the ensemble may capture a comprehensive aspects of  forensic statistics of  the unaltered images and enforce the generator to only produce anti-forensically attacked images that reside on the overlapping areas of  decision boundaries of unaltered classes.  Our intuition that  the ensemble can improve the transferability of the attack is that if the generated anti-forensically images can fool each surrogate CNN in the ensemble,  they most likely can fool other CNNs not used during training the attack. 

To improve the transferbility of the proposed anti-forensic attack,  it is critical to ensure the diversity of surrogate CNNs in the ensemble.  We propose to increase the diversity by varying CNN architectures and definitions of classes.   Particularly,  the attacker can choose multiple CNN architectures,  then for each CNN architecture,  the attacker can obtain multiple versions of CNNs  by changing the definition of classes for manipulation detection,  manipulation identification,  and manipulation parameterization,  as described in Section~\ref{sec: CNNs}.

In the perfect knowledge scenario,  since the attacker has full access to the victim CNN possessed by the investigator,   the victim CNN can be directly used as one surrogate CNN to train the proposed anti-forensic attack.   While the attacker does not need to choose other architectures,  empirically we found that using an ensemble with various  class definitions of the victim CNN can improve the performance of the anti-forensic attack even in the perfect knowledge scenario.  These findings will be demonstrated  in Section~\ref{subsec: perfect} via experiments.
In limited knowledge scenarios,  since the attacker has no direct access to the victim CNN possessed by the investigator,  each surrogate CNN should be trained using CNN architectures of the attacker's choice and training data collected by the attacker.  The attacker should train surrogate CNNs with varying architectures and class definitions to ensure the diversity of the ensemble.

\vspace{-1em}
\subsection{ Creation of Training Data}
\label{sub: p2p}
\subhead{Pixel to Pixel Correspondence} We found that the transferabiliy of the proposed anti-forensic GAN attack heavily depend on the strategy of creating training data which will be demonstrated in later sections via experiments.  We propose that  training data  should be created as a pair of the  unaltered image and its corresponding manipulated image.  
Specifically,  each pixel in the unaltered image should have a corresponding pixel in the manipulated image.   One explanation for this finding is that  the pixel  to pixel correspondence between the unaltered image and the manipulated image  can enforce the proposed attack to ignore difference caused by image contents and only learn the difference resulted from forensic information.   Since the discriminator and surrogate CNNs in the proposed framework take in images of the same size,   manipulations that distort the shape of images,  such as resizing or barrel distortion,  may decrease the transferability of the proposed anti-forensic GAN attack.

\vspace{-1em}
\subsection{Loss functions}
To train the proposed anti-forensic GAN attack,  we now define loss functions for the generator and the discriminator.

\subhead{Generator's Loss}
This loss function is formulated to ensure generated anti-forensically attacked images maintain high visual quality,  and  can fool the ensemble of surrogate CNNs and the discriminator.   The generator's loss $\mathcal{L}_{G}$ is formulated as the weighted sum of three terms,  perceptual loss $\mathcal{L}_{p}$,  classification loss $\mathcal{L}_{c}$ and adversarial loss $\mathcal{L}_{a}$, 

\begin{equation}
	\label{eq:g_loss}
	\mathcal{L}_{G}=\alpha\mathcal{L}_{p}+ \sum_{s=1}^{S}\beta^{(s)}\mathcal{L}_{c}^{(s)}+ \gamma\mathcal{L}_{a},
	\vspace{-0.3em}
\end{equation}
where $S$ represents  the number of surrogate CNNs in the ensemble and $\alpha$,  $\beta^{(s)}$,  $\gamma$ are weights.

\underline{\smash{Perceptual loss}} is used to ensure the visual quality of anti-forensically attacked images produce by the proposed attack.  It is formulated as the mean absolute difference of pixel values between the unaltered image and its corresponding generated anti-forensically attacked image.  For an unaltered image $I$ of size $w \times h$ and its corresponding manipulated image $I'$,  perceptual loss is expressed as,
\begin{equation}
	\mathcal{L}_{p}=\frac{1}{w\times h}\sum_{i=1}^{w}\sum_{j=1}^{h}\mid I_{i,j}-G(I')_{i,j}\mid,
	\label{eq: perceptual_loss]}
\end{equation}
where subscripts $i, j$ represent the pixel's coordinates.

We note that when training the attack to remove traces left by manipulations that distort the image's shape,  formula~\ref{eq: perceptual_loss]} should be modified and  calculated between the generated anti-forensically attacked image and the manipulated image $I'$ rather than the unaltered image $I$.  However,  this modification may degrade the transferability of the attack. 

\underline{\smash{Classification losses}} are used to ensure the anti-forensically attacked image produced by the proposed attack can  fool the ensemble of surrogate CNNs.   For $s^{th}$ surrogate CNN of the ensemble,   let $C(\cdot)_s$ represent this surrogate CNN's softmax output,  then the classification loss $\mathcal{L}_{c}^{(s)}$ pertaining to the  $s^{th}$ surrogate CNN is formulated as the softmax cross-entropy between the generated anti-forensically attacked image and the unaltered class,
\begin{equation}
	\mathcal{L}_{c}^{(s)}=-\sum_{k=1}^{K} t_k \log\left( C_s(G(I'))_k\right),
	\label{eq: classification_loss}
\end{equation}
where $K$ is the number of the classes,  $t_k$ is the $k^{th}$ entry of ideal softmax vector with  a $1$ at the location of the unaltered class and $0$'s elsewhere. 

\underline{\smash{Adversarial loss}} is used to ensure the anti-forensically attacked image produced by our proposed attack can fool the discriminator.  Adversarial loss is formulated as the sigmoid cross-entropy between the discriminator's output of the generated anti-forensically attacked image and $1$.
\begin{equation}
	\mathcal{L}_{a}=\log(1- D(G(I'))),
	\label{eq: adversarial_loss}
\end{equation} 

\subhead{Discriminator's Loss}
The loss function is formulated to ensure the discriminator can distinguish between unaltered images and generated anti-forensically attacked images.   The discriminator's loss $\mathcal{L}_D$  is expressed as,
\begin{equation}
	\mathcal{L}_D=\log D(I)+\log(1- D(G(I')))
	\label{eq: discriminator_loss}
\end{equation}

\vspace{-1em}
\subsection{ Deployment of the Proposed Attack}
After training the proposed anti-forensic GAN framework,  the attacker can use the generator to anti-forensically remove the forensic traces left by manipulation operations from forged images.  The trained generator can be applied to color images of any size.  If the attacker does not possess enough computational resources to attack full-size images at once,   the attacker can divide the full-size color image into smaller patches and use the generator to attack each image patch individually,  and then group the anti-forensically attacked  patches together in the original order to form the full-size attacked image.

%% file: sections/eval_metrics_v2_MCS.tex
\label{sec: metrics}
The proposed anti-forensic GAN attack should be able to fool victim CNNs,   and leaves no visible distortions.  

\subhead{Attack Success Rate (ASR)} To evaluate the performance of the anti-forensic GAN attack on fooling the victim CNN,  we defined attack success rate as the percentage of the generated anti-forensically attacked image patches are classified to the ``unaltered class".

\subhead{Mean SSIM and Mean PSNR} The anti-forensically attacked images should not contain any perceptible distortions.  To evaluate the visual quality of anti-forensically attacked images produced by our attack,  we calculate the average of SSIM and PSNR between generated the anti-forensically attacked images and corresponding manipulated images.  

%% file: sections/experiment_setup_v2_MCS.tex

We conducted a series of experiments to evaluate the performance of our proposed anti-forensic GAN attack against a variety of manipulation forensic CNNs built under each scenario described in Section~\ref{sec: scenarios}.  We assume that the investigator will use CNNs trained for manipulation detection,  manipulation identification, or manipulation parameterization to authenticate images,  and will trust the results of their CNNs.   

To show that our proposed attack is a general attack and can remove forensic traces left by varying manipulation operations,  we selected three common manipulations,  and for each manipulation we chose five parameters that cover a reasonable range of values.  The chosen manipulations and parameters are shown in Table~\ref{tab: manipulations}.  
These manipulations were chosen for several reasons.  First, these manipulations can be easily parameterized and used to train manipulation parameterization classifiers.  Second, these manipulations  do not alter the shape or size of an image, thus making it straightforward to create training data using the pixel to pixel correspondence rule.  
Additionally, finite computational resources placed practical limitations on the number of manipulations and associated manipulation parameters that we could run in our extensive experiments.  
%
Specifically, it took $3,000$ computational hours to run the experiments in this paper with an Nvidia 1080Ti GPU.  
The number of manipulations and associated manipulation parameters could be increased given greater computational power and larger storage, however this was not feasible given our current computational resources.


\begin{table}[t]
\centering
 	\caption{Manipulation operations and their associated parameters. }
 	\begin{tabular}{|l c|}
 		\hline
 		\textbf{Manipulations} & \textbf{Parameters}\\\hline
 		Additive Gaussian white noise &$\mu=0,\sigma=0.5,1,1.5,2,2.5$\\\hline
 		Gaussian blurring  & $\sigma=1,1.5,2,2.5,3,3.5,4,4.5$\\\hline
 		Median filtering &  filter size $=3,5,7,9,11$\\\hline
 	\end{tabular}
 	\label{tab: manipulations}
 \end{table}
 
\vspace{-1em}
\subsection{Data Preparation}
We started with preparing data for the experiments.   We used the Dresden image database~\cite{Dresden},  and our  data were created from  $169,620$ full-size JPEG images taken by 70 unique devices  of 27 camera models.   First we randomly selected $10\%$ of  images as the evaluation set,  \textit{``eval-set"}.  The  \textit{``eval-set"}  were only used for evaluating the performance of forensic CNNs ``before and after" the proposed attack.   Since under all limited knowledge scenarios described in Section~\ref{sec: scenarios} we assume the attacker has no access to the investigator's training data,  we built two disjoint training data sets by randomly bisecting the rest $90\%$ of images,  \textit{``I-set"} for the investigator  and  \textit{``A-set"} for the attacker.  As a result,  we ensure that \textit{``eval-set"},   \textit{``I-set"} and  \textit{``A-set} contains different images,  and shares no same statistics.     \textit{``I-set"}  were used as by the investigator to train manipulation forensic CNNs  for authenticating images.   \textit{``I-set"} can also be used by the attacker only under the perfect knowledge scenario.   \textit{``A-set} were used by the attacker to  train the surrogate CNNs and then the proposed anti-forensic GAN attack under all limited knowledge scenarios.  Next,  for each data set we created manipulation images  by applying each combination of manipulation and parameter shown in Table~\ref{tab: manipulations} to every single full-size image.   The manipulated images were saved as PNG files.  Thus,  each data set contains $15$ unique classes of manipulated images and one class of unaltered images.  Then we divided full-size images into $256 \times 256$ non-overlapping image patches.  The following experiments were evaluated on image patches.  

\vspace{-1em}
\subsection{Baseline Performance of Manipulation Forensic CNNs}
\label{subsec: CNN_performance}
In this experiment,  we  evaluated the baseline classification accuracies of CNNs trained on \textit{``I-set"} and \textit{``A-set"}.   CNNs trained on both data sets were evaluated on  \textit{``eval-set} for fair comparison.  We selected six CNN architectures as they can achieve the state-of-art performance on manipulation forensics.  These CNN architectures are MISLnet~\cite{MISLNet},  SRNet\cite{SRNet},  PHNet~\cite{PHnet},  TransferNet~\cite{TransferNet},  DenseNet\textunderscore BC~\cite{DenseNet},  and  VGG-19~\cite{VGGnet}.  We note that to train TransferNet CNNs on color images,  we modify the high pass filter layer of TransferNet from one high pass filter to three identical high pass filters. 
For each architecture,  we trained CNNs using the three class definitions.   We grouped the images patches to make two classes (one unaltered class vs.  one manipulated class) for the manipulated detection,  four classes (one unaltered class vs.  three manipulated classes) for the manipulation identification,  and 16 classes  (one unaltered class vs. 15 manipulated classes) for the manipulation parameterization.   

Due to the difference in goals and conditions of experiments,  we chose the best set of hyperparemeters for each CNN architecture if yields the highest classification accuracy after grid search.  CNNs of same CNN architecture and  different class definitions were trained using the same hyperparemeters.   All CNNs were trained from scratch for $43$ epochs,  and would stop early if the training loss started to increase.  Weights were initialized using Xavier initializer~\cite{Xavier} and biases were initialized as $0$'s.  Weights and biases were optimized using stochastic gradient descent.   For TransferNet,  the batch size was $50$,  and  the learning rate started with $0.001$ and decayed $10\%$ every $5,000$ iterations.  For other five CNN architectures,  batch size was $25$,  and the learning rate started from $0.0005$ and decayed half every $4$ epochs.    

CNNs trained on  \textit{``I-set"} are the victim CNNs that attacker attempts to fool,  and the investigator uses them to authenticate images.  The  baseline classification accuracies of  CNNs trained on  \textit{``I-set"}  are shown in Table~\ref{tab: I-set}.    On average,   for \textit{``I-set"},  we achieved the classifcation accuracy of $99.29\%$ for manipulation detection,  $98.51\%$ for manipulation identification,  and $77.93\% $ for manipulation parameterization.  
CNNs trained on \textit{``A-set"} are the surrogate CNNs  used by the attacker  to built the ensemble for  training the proposed anti-forensic GAN attack under limited knowledge scenarios.    The  baseline classification accuracies of  CNNs trained on  \textit{``A-set"}  are shown in Table~\ref{tab: A-set}.  On average,  we achieved the classification accuracy of $99.42\%$  for  manipulation detection,  $98.39\%$ for manipulation identification,  and $79.08\%$ for  manipulation parameterization.   These results shows that  CNNs used in our experiments were well trained  and consistent with the state-of-art  performances on manipulation forensics.    Furthermore,  the differences between classification accuracies of CNNs trained on  \textit{``I-set"} and \textit{``A-set"} indicate CNNs possessed by the investigator and the attacker are comparable but not identical.

\begin{table}[t!]
\centering
\caption{\textit{I-set},  baseline classification accuracies for six CNN architectures and three class definitions.}
\label{tab: I-set}
\begin{tabular}{|l |ccc|}
\hline
\textbf{CNN Architect.} & \textbf{Detection} & \textbf{Classification}&\textbf{Parameterization}\\\hline
MISLnet& 99.84\% & 99.55\% & 86.24\% \\
TransferNet& 99.20\%& 98.04\%& 65.27\% \\
PHNet &99.58\% & 98.94\%&86.58\% \\
SRNet & 99.16\%& 99.36\%  & 81.30\% \\
DenseNet\textunderscore BC &98.13\%&95.66\%&65.50\%\\
VGG-19& 99.87\% & 99.50\%&82.67\%\\
\textbf{Avg.}& \textbf{99.29\%}&\textbf{98.51\%}&\textbf{77.93\%} \\\hline
\end{tabular}
 \end{table}

 \begin{table}[t!]
\centering
\caption{\textit{A-set},  baseline classification accuracies for fsix CNN architectures and three class definitions.}
\label{tab: A-set}
\begin{tabular}{|l |ccc|}
\hline
\textbf{CNN Architect.} & \textbf{Detection} & \textbf{Classification}&\textbf{Parameterization}\\\hline
MISLnet &99.12\% & 99.08\% & 87.44\%\\
TransferNet& 99.54\%& 98.66\%& 67.81\% \\
PHNet & 99.58\%& 98.76\%&84.79\%\\
SRNet & 99.47\%&  98.50\%& 83.78\%\\
DenseNet\textunderscore BC &98.89\%& 95.91\%&68.60\%\\
VGG-19& 99.90\% &99.41\% &82.11\%\\
\textbf{Avg.}&\textbf{99.42\%} &  \textbf{98.39\%} & \textbf{79.08\%}\\\hline
\end{tabular}
 \end{table}
 
\vspace{-1em}
\subsection{Training Proposed Anti-Forensic GAN}

All attacks demonstrated in this paper were trained from scratch for $12$ epochs. Weights for each loss term were all $1's$.  Weights of the generator and the discriminators are initialized using Xavier initializer~\cite{Xavier} and biases were initialized as $0$'s.  The generator was optimized using Adam optimizer~\cite{kingma2017adam} and the discriminator was optimized using stochastic gradient descent.  The learning rate was started with $0.0001$ and decayed half every $4$ epochs.

%% file: sections/perfect.tex
\vspace{-1em}
\subsection{Perfect Knowledge}
\label{subsec: perfect}

\begin{table}[t!]
\centering
\caption{Attack success rates (ASRs) achieved by different attack strategies in the perfect knowledge scenario.}
\label{tab: perfect}
\resizebox{0.5\textwidth}{!}{
	\begin{tabular}{|l cccc|}
		\hline
		  \multicolumn{5}{|c|}{\textbf{Proposed Anti-Forensic GAN}} \\
		\hline
		\textbf{CNN Architect.} & \textbf{Detection} & \textbf{Classification}  &\textbf{ Parameterization}& \textbf{Avg.}\\
		MISLnet& 0.98& 0.98& 0.98&0.98\\
		TransferNet &1.00 &0.99&1.00&1.00 \\
		PHNet &0.84 & 0.99 &0.99&0.94\\
		SRNet &0.96 & 0.99&0.97&0.97\\
		DenseNet\textunderscore BC & 0.99 &0.96 &0.98&0.98\\
		VGG-19 & 0.99 &0.96 &0.98&0.98\\
		\textcolor{blue}{\textbf{Avg.} }& \textcolor{blue}{\textbf{0.96}} & \textcolor{blue}{\textbf{0.98}} & \textcolor{blue}{ \textbf{ 0.99}} & \textcolor{blue}{\textbf{0.98}}\\
       \hline
	  \multicolumn{5}{|c|}{\textbf{Removing Discriminator}} \\
	    	\hline
         \textbf{CNN Architect.} & \textbf{Detection} & \textbf{Classification}  &\textbf{ Parameterization}& \textbf{Avg.}\\
         MISLnet&0.90&0.97&1.00&0.95\\
         TransferNet &1.00 &1.00&1.00&1.00\\
         PHNet&0.78&0.93&0.98&0.89\\
		SRNet&0.53&0.93&0.98&0.81\\
		DenseNet\textunderscore BC & 0.89&1.00 &1.00 &0.96\\
		VGG-19 &0.87 & 0.99 &0.99 &0.95\\
		\textbf{Avg.} & \textbf{0.83} &\textbf{0.97}&\textbf{0.99} &\textbf{0.92} \\
	    	\hline
  \multicolumn{5}{|c|}{\textbf{MISLGAN~\cite{Mislgan}}} \\
	    	\hline
         \textbf{CNN Architect.} & \textbf{Detection} & \textbf{Classification}  &\textbf{ Parameterization}& \textbf{Avg.}\\
     MISLnet& 0.55 & 0.95& 0.84 &0.78\\
	TransferNet & 0.99 & 1.00 & 1.00&1.00 \\
	PHNet&0.90 & 0.97&0.94&0.94\\
	SRNet & 0.88& 0.90 & 0.82&0.87\\
	DenseNet &0.90&0.94&0.94&0.93\\
	VGG-19 & 0.71 &0.97&0.96&0.88\\
		\textbf{Avg.} &  \textbf{0.82} & \textbf{0.96} & \textbf{0.92}& \textbf{0.90}\\
		\hline
          \multicolumn{5}{|c|}{\textbf{Standard GAN}} \\
         \hline
         \textbf{CNN Architect.} & \textbf{Detection} & \textbf{Classification}  &\textbf{Parameterization}& \textbf{Avg.}\\
         MISLnet&0.08&0.01 &0.00&0.03\\
         TransferNet & 0.02&0.00 &0.00 &0.01\\
         PHNet&0.06 &0.00&0.00&0.02\\
		SRNet & 0.04&0.20 &0.18&0.14\\
		DenseNet\textunderscore BC &  0.08 &  0.24& 0.26 &0.19\\
		VGG-19 &0.05  &  0.75 & 0.57 &0.46\\
		\textbf{Avg.} &\textbf{0.06}& \textbf{0.20} & \textbf{0.17} &\textbf{0.14} \\
        \hline
        	\end{tabular}
	}
\end{table}
In this experiment,  we evaluated the baseline performance of the proposed anti-forensic GAN attack in the perfect knolwedge scenario.  In this scenario,  we assume the attacker has equal amount of information as the investigator (i.e whatever Bob has access to,  Alice has the equal access).   Specifically,  the attacker has access to the investigator's  training data.  The attacker can either obtain investigator's trained CNN or obtain an identical copy.   Also the attacker knows about the architecture and class definitions of the investigator's  CNN.     

To launch the attack on the victim CNN using the proposed anti-forensic GAN attack,  the attacker should first obtain an ensemble of surrogate CNNs and then train the proposed anti-forensic GAN attack by integrating the ensemble of surrogate CNNs into the training  phase.   Since the attacker has  access to  the trained victim CNN,  the attacker has an advantage including the victim CNN into the ensemble.   Furthermore,  since the attacker knows about the architecture of the victim CNN,  an easy approach to make the ensemble is to adopt the CNNs of  same  architecture and other class definitions.  As a result,  the ensemble of surrogate CNNs contains the victim CNN and the other two CNNs of the same architecture but different class definitions.   Since the attacker also has access to the investigator's training data,   the attacker could use  \textit{``I-set''} to build training data for the proposed attack to avoid statistic discrepancy caused by mismatching training data.    We note that the  pixel to pixel correspondence in creating the training data  is not required in the perfect knowledge scenario for our proposed attack.  However we found that it can drastically improve the performance of other attack strategies used for comparison in this subsection.   To make a  fair comparison,  the creation of training data follows the rules of pixel to pixel correspondence in this evaluation.  After training,  the trained generator can be used to  attack the investigator's CNN of the same architecture and any class definition. 

To demonstrate this,   we treated each CNN demonstrated trained on \textit{``I-set"}  as a victim CNN to fool.    For each of the six CNN architectures,   an ensemble of the surrogate CNNs was made by grouping the manipulation detector,  the manipulation identifier and the manipulation parameterizer of this architecture.  Then we trained the proposed anti-forensic GAN attack using the ensemble.   Hence,  we only trained the proposed attack once for CNNs of the same architecture.  
The trained generator then would be used to attack manipulated image patches and produced anti-forensically attacked image patches.   All anti-forensically attacked image patches were saved to disk as PNG files and then read back for classification.  This is to ensure the pixel values of generated anti-forensic attacked images reside on the legit range from 0 to 255.   Next,  we evaluated the performance of the proposed attack by classifying the anti-forensic attacked images using each victim CNN.  

We evaluated the our proposed attack on $45,000$ manipulated image patches from \textit{``eval-set "}.  The attack success rates for fooling each victim CNN  are demonstrated in Table~\ref{tab: perfect}.  On average,  our proposed anti-forensic GAN attack achieved the attack success rate of $0.96$  for manipulation detection,  $0.98$  for manipulation classification,  and $0.99$ for manipulation parameterization.  The mean of attack success rates regardless of the class definitions was $0.98$.   It means that the proposed anti-forensic GAN attack can successfully full forensic CNNs in the perfect knowledge scenario. 

\subhead{The Effect of the Discriminator}
We explored other generative attack strategies,  and demonstrated the performance of the trained generator when training without a discriminator.  Removing the discriminator from the proposed attack,  the generate will  only learn the forensic information provided by the ensemble of surrogate CNNs,  and will not compete with the discriminator.   From Table~\ref{tab: perfect},  we see while on average the attack success rates of this attack strategy were comparable with our proposed anti-forensic GAN attack in terms of manipulation classification and manipulation parameterization,  the attack success rate for manipulation detection was $13\%$ lower than our proposed attack.  The difference was especially significant for manipulation detection classifier of SRNet architecture.  Removing the discriminator,  the attack success rate dropped $43\%$.  The results show that discriminator is needed to improve the attack. 

\subhead{The Effect of the Ensemble of Surrogate CNNs}
We compared the performance of our proposed attack with MISLGAN attack~\cite{Mislgan}.   While MISLGAN was initially proposed as a white-box attack for camera model falsification,  it can be easily adapted to anti-forensically remove traces left by manipulations.  To make a MISLGAN attack,  we integrated the victim CNN to train the generator,   instead of the ensemble.  The results are shown in Table~\ref{tab: perfect}.  On average,  our proposed anti-forensic GAN attack achieved $14\%$ higher attack success rate for manipulation detection,  $2\%$ higher for manipulation classification and $7\%$ higher for manipulation parameterization.  Generally,  our proposed anti-forensic GAN attack achieved $8\%$ higher attack success rate than MISLGAN.   This results shows that introducing the ensemble of surrogates CNNs is very critical for improving the performance of the  attack,  even in the perfect knowledge scenario.

\subhead{Comparing with Standard GANs} We compared with standard GAN framework.  Since the attacked knows the architecture of the investigator's CNN,  the attacker can modify the last layer of each CNN architecture to be a single node and construct a discriminator.  Combined with the generator,   we trained a standard GAN attack and computed attack success rate for each victim CNN.   The results show that standard GANs achieved the lowest attack success rate among all attack strategies we demonstrated in Table~\ref{tab: perfect}.  It indicates that the standard GANs  are not suitable to falsify forensic traces,  and integrating the ensemble of surrogates can solve this problem.

\subhead{Visual Quality and Inspection} We evaluated the visual quality of generated anti-forensically attacked images by computing the mean PSNR and the mean SSIM between manipulated image patches and generated anti-forensically attacked image patches.  The mean PSNR  was $54.64$ and the mean SSIM was  $0.9986$.  The results show that in the perfect knowledge scenario,  our proposed  attack will not introduce perceptible distortion to the anti-forensically attacked images. 


%% file: sections/data_mismatch.tex
\vspace{-1em}
\subsection{Training Data Mismatch}
\label{subsec: data_mismatch}

In this experiment,  we evaluated the performance of the proposed anti-forensic GAN attack in a limited knowledge scenario when the training data of the attacker are different from the investigator.  In this scenario,   we assume he attacker has no access to the investigator's trained forensic CNN and  training data.  We also assume the attacker knows about the architecture of investigator's CNN.   As a result,  the attacker cannot use  the victim CNN or an identically trained copy of the victim CNN to train the attack. 
The anti-forensically attacked images produced by the attacker should  transfer across training data set to fool the investigator's CNN. 

To launch the attack on the victim CNN using our proposed anti-forensic GAN attack,  the attacker should first form the ensemble of surrogate CNNs of different architectures and class definitions.   Since the attacker knows about the architecture of the victim CNN,  the attacker could  group the surrogate CNNs of  this architecture plus other architectures to form an ensemble.  Next,  the attack can train the proposed anti-forensic GAN  using the ensemble of surrogate CNNs.   Since the attacker has no access to the investigator's training data,   the attacker should create training data for the attack from  \textit{``A-set"}.   The creation of training data follows the rules of pixel to pixel correspondence.   

To demonstrate it,  we treated each CNN trained on \textit{``I-set"} as a victim CNN to fool.   We trained the proposed attack using an ensemble of  surrogate CNNs trained on \textit{``A-set"}.  Ideally we could  use all surrogate CNNs to form the ensemble to capture the most comprehensive forensic information of unaltered images,  and thus we would only train the proposed attack once for all victim CNNs.   However,   in practice due to limitation of computer memory,   we could not load  all surrogates CNNs into the memory to train one single generator.  Since  VGG-19 CNNs take the most computer memories,  we trained one attack against VGG-19 CNNs,  and one attack against all other CNNs.   The ensemble of surrogate CNNs built for attacking VGG-19 consists of three surrogate CNNs of VGG-19 and three surrogate CNNs of MISLnet.  The ensemble of surrogate CNNs built for attacking other victim CNNs consists of all other surrogates CNN excluding VGG-19 CNNs.  
After training the attack,  we used the trained generator to attack each manipulation image patches and produce anti-foresically attacked image patches.  We evaluated the performance of our proposed  attack by classifying the generated anti-forensic attacked  images using each victim CNN.   

We evaluated the performance of our proposed anti-forensic attack on $45,000$ manipulated image patches from \textit{``eval-set''}.  The attack success rates for fooling each victim CNNs are shown in Table~\ref{tab: data_mismatch}.  Except for manipulation paramerization of DenseNet\textunderscore BC architecture,  our attack achieved high attack success rates on the victim CNNs.  On average,  our proposed anti-forensic GAN attack achieved the attack success rate of $0.98$ for manipulation detection,  $0.85$  for manipulation classifcation,  and $0.87$ for manipulation parameterization.   The mean of attack success rates regardless of the class definition was  $0.90$.   The results show that the proposed anti-forensic GAN attack can strongly fool forensic CNNs trained on different training data. 



 \begin{table}[t]
	\centering
	\caption{Attack success rates (ASRs) achieved by different attack strategies in training data mismatch scenario. }
	\label{tab: data_mismatch}
	\resizebox{0.50\textwidth}{!}{
		\begin{tabular}{|l cccc|}
				\hline
			\multicolumn{5}{|c|}{\textbf{Proposed Anti-Forensic GAN}}\\
			    \hline
				\textbf{CNN Architect.} & \textbf{Detection} & \textbf{Classification}  &\textbf{ Parameterization} & \textbf{Avg.}\\
				MISLnet &1.00& 0.87 & 1.00 &0.96\\
				TransferNet &1.00& 0.99& 0.98&0.99\\
				PHNet&0.98& 1.00 & 0.96&0.98\\
				SRNet &0.93 &0.97 &0.78&0.89\\
		 	   DenseNet\textunderscore BC & 0.99 &0.31 & 0.64&0.65\\
		   	   VGG-19 & 0.98  &0.95 & 0.86&0.93\\
			\textcolor{blue}{\textbf{Avg. }} & \textcolor{blue}{\textbf{0.98}} & \textcolor{blue}{\textbf{0.85}}& \textcolor{blue}{\textbf{0.87}}&\textcolor{blue}{\textbf{0.90}}\\
	     \hline
  \multicolumn{5}{|c|}{\textbf{MISLGAN~\cite{Mislgan}}}\\\hline
	     \textbf{CNN Architect.} & \textbf{Detection} & \textbf{Classification}  &\textbf{ Parameterization} & \textbf{Avg.}\\
	     	MISLnet &  0.78 &0.25  &  0.91&0.64\\
			TransferNet  &  1.00 &1.00 &1.00 &1.00\\
			PHNet &  0.62 &0.00 &0.19 &0.27\\
			SRNet  &  0.46 & 0.06& 0.34&0.29\\
		 	DenseNet\textunderscore BC  & 0.21  &0.08 &0.01 &0.10\\
		   	VGG-19 & 0.10  & 0.53& 0.52&0.38\\
			\textbf{Avg. }  &  \textbf{0.53} &\textbf{0.32}&\textbf{0.50} &\textbf{0.45}\\
			    \hline
	      \multicolumn{5}{|c|}{\textbf{Removing Architecture Diversity}}\\
			    \hline
	\textbf{CNN Architect.} & \textbf{Detection} & \textbf{Classification}  &\textbf{ Parameterization} & \textbf{Avg.}\\
				MISLnet & 0.79& 0.17 & 0.99& 0.65\\
				TransferNet & 0.99& 1.00 & 0.88 & 0.95\\
				PHNet& 0.97& 0.68 &0.83 & 0.82\\
				SRNet & 0.53&0.21 & 0.01& 0.25\\
		 	   DenseNet\textunderscore BC & 0.70 & 0.12 & 0.71 &0.51\\
		   	   VGG-19 & 0.98& 0.97& 0.65 & 0.87\\
		   	   \textbf{Avg.}& \textbf{0.83}&\textbf{0.53} &\textbf{0.68} & \textbf{0.68}\\
		   	   \hline
		   	      \multicolumn{5}{|c|}{\textbf{Training without Pixel to Pixel Correspondence}}\\
			    \hline
				\textbf{CNN Architect.} & \textbf{Detection} & \textbf{Classification}  &\textbf{ Parameterization} & \textbf{Avg.}\\
				MISLnet & 0.95& 0.71 &0.88 &0.85\\
				TransferNet  & 1.00   & 0.98&0.98 &0.96\\
				PHNet &  0.97 & 1.00& 0.88&0.95\\
				SRNet  &  0.93 &0.92 &0.73 &0.86\\
		 	   DenseNet\textunderscore BC  &  0.70 &0.01 & 0.13&0.28 \\
		   	   VGG-19  &  0.90 & 0.79 &0.51 & 0.73\\
		   	   \textbf{Avg.}  &  \textbf{0.91} & \textbf{0.74}&\textbf{0.69} &\textbf{0.78} \\\hline
			\end{tabular}
		}
\end{table}

\subhead{Comparing with MISLGAN} We compared the performance of the proposed attack with the state-of-art GAN-based atttack,  MISLGAN~\cite{Mislgan}.  To attack each victim CNN,  we trained the MISLGAN using the surrogate CNN built with the same architecture and class definition,  and tested the performance of anti-forensically attacked images using the victim CNN.    The results are shown as in Table~\ref{tab: data_mismatch}.  Comparing with MISLGAN attack, the average  attack success rates achieved by our proposed attack almost doubled for all class definitions.   Except for TranferNet CNNs,  the attack success rates for other CNNs all dropped significantly comparing to performance of the MISLGAN attack in the perfect scenario shown in Table~\ref{tab: perfect}.   It may indicate the TransferNet CNNs are not robust to adversarial attacks.   The results are consistent with previous finding that  the  attack on forensic CNNs  can hardly transfer across training data~\cite{transferability_Barni},  and show  that  the ensemble of surrogate CNNs is important  to make the attack transfer.

\subhead{The Effect of CNN Architecture Diversity } We conducted experiments to show that building the ensemble of surrogate CNN with diverse architecture is critical for the attack to  achieve transferability in limited knowledge scenarios.   To demonstrate this,  for each victim CNN,  we used three surrogates CNNs of same architecture as the victim CNN to build the ensemble,  and then trained the attack.  Shown in Table~\ref{tab: data_mismatch},  the attack success rates achieved when only using one CNN architecture significantly decreased.  On average,  the attack success rate dropped $15\%$ for manipulation detection,  $32\%$ for manipulation classification,  and $19\%$ for manipulation parameterization.   Particularly,  attack success rates for SRNet show the most significant differences when training with diverse CNN architectures in the ensemble,  the attack success rate increased $40\%$ for manipulation detection,  $75\%$ for manipulation  classification,  and $77\%$ for manipulation parameterization.  These results show the training the attack with diverse CNN architectures can improve the transferability of the attack dramatically.  

\subhead{The Effect of Pixel to Pixel Correspondence} We compared with the attack success rates when training data was not created following the rule of pixel to pixel correspondence. Training data were created from \textit{``A-set"},  but we do not require correspondence between pixels for each training batch.   The visual quality loss was modified to compute between the anti-forensically attacked image patch output by the  generator and the manipulated image patch input to the generator.  The results are shown in Table~\ref{tab: data_mismatch}.  Without using pixel to pixel correspondence,  on average,  the attack success rates dropped $7\%$ for manipulation detection,  $11\%$  for manipulation classification and $19\%$ for manipulation classifiers.  The results show that using the pixel to pixel correspondence can help improve the transferability of the attack.

\subhead{Visual Quality} We evaluated the image quality of the anti-forensically attacked images produces by our proposed attack  by calculating the mean PSNR and the mean SSIM between  manipulated image patches and generated anti-forensically attacked image patches.  The mean PSNR is $47.12$ and the mean SSIM is $0.9944$.  The results show that in the training data mismatch scenario,  our proposed  attack will not introduce perceptible distortion to the anti-forensically attacked images.  


%% file: sections/para_mismatch_v2_MCS.tex
\vspace{-1em}
\subsection{Training Data and Manipulation Parameter Mismatch}
\label{subsec: para_mismatch}
\begin{table}[t!]
\centering
\caption{Classification accuracies of surrogate CNNs  for building the ensemble in the training data and manipulation parameter mismatch scenario. }
\label{tab: accuracy_3m4p}
\begin{tabular}{|l|ccc|}
	\hline
 \textbf{CNN Architect.} & \textbf{Detection} & \textbf{Classfication}  &\textbf{Parameterization} \\\hline
MISLNet& 99.02\%& 99.28\%&90.88\%\\
Zhan et. al &99.27\%&97.99\%&68.13\%\\
PHNet&99.44\%&98.86\%&90.44\%\\
SRNet&99.65\%&98.72\%& 88.31\%\\
DenseNet\textunderscore BC &94.80\%& 96.58\%& 73.64\%\\
VGG-19 & 99.97\% & 99.26\% & 88.09\%\\
\textbf{Avg.} & \textbf{98.69\%}& \textbf{98.45\%} &\textbf{83.25\%}\\\hline
\end{tabular}
\end{table}

\begin{table}[t!]
\centering
\caption{Attack success rates (ASRs) achieved by the proposed anti-forensic GAN attack in the training data and manipulation parameter mismatch scenario.  }
\label{tab: para_mismatch}
\resizebox{0.50\textwidth}{!}{
	\begin{tabular}{|l cccc|}
\hline
		\multicolumn{5}{|c|}{\textbf{ All Manipulation Parameters}}\\
\hline
		\textbf{CNN Architect.} & \textbf{Detection} & \textbf{Classification}  &\textbf{ Parameterization}  & \textbf{Avg.}\\
		MISLnet & 0.99& 0.53 & 0.35 & 0.62 \\
		TransferNet & 1.00 &0.97 &0.94 & 0.97\\
		PHNet &  0.96 &1.00 &0.99&0.98 \\
		SRNet & 0.59 & 0.90 & 0.52 &0.68 \\
		DenseNet\textunderscore BC & 0.99 & 0.49 & 0.21 & 0.56 \\
		VGG-19 & 0.83 &0.59 &0.74 &0.72\\
		\textcolor{blue}{\textbf{Avg.}} & \textcolor{blue}{ \textbf{0.89}} & \textcolor{blue}{\textbf{0.75}} & \textcolor{blue} {\textbf{0.63}} &\textcolor{blue}{ \textbf{0.76}}\\
		\hline
		\multicolumn{5}{|c|}{\textbf{Unseen  Manipulation Parameters Only}}\\
\hline
		\textbf{CNN Architect.} & \textbf{Detection} & \textbf{Classification}  &\textbf{Parameterization}  & \textbf{Avg.}\\
		MISLnet & 1.00 & 0.51  & 0.28 &0.60 \\
		TransferNet & 1.00 & 0.98 & 0.99 &0.99\\
		PHNet & 0.92  & 1.00 &0.95&0.96 \\
		SRNet  & 0.60 &0.95 &0.49 &0.68\\
		DenseNet\textunderscore BC & 0.99 &0.35 & 0.20 &0.51\\
		VGG-19 & 0.75  &0.51 & 0.79 &0.68\\
		\textcolor{blue}{\textbf{Avg.}} & \textcolor{blue}{ \textbf{0.87}} & \textcolor{blue}{\textbf{0.72}} &\textcolor{blue}{ \textbf{0.62}}&\textcolor{blue}{\textbf{0.74}}\\
		\hline
		\end{tabular}
}
\end{table}

In this experiment, we evaluated the performance of the proposed anti-forensic GAN attack in a limited knowledge scenario when the attacker has partial knowledge about the manipulations of the investigator's interest.  In this scenario,  we assume the attacker has no access to the investigator's trained CNNs and  training data.  We also assume the attacker knows about architecture of the investigator's CNN and the manipulations  the investigator tries to identify.  However,  the attacker has no perfect knowledge of the parameters associated with each manipulation.  Specifically,  the attacker only gather a subset of the manipulation parameters to create the attacker's training data.     Moreover,  since the attacker does not know all manipulation parameters,  it  will cause a more severe statistical discrepancy on the attacker's CNNs comparing with the training data mismatch scenario demonstrated earlier.    Particularly,  the investigator's CNNs can output more classes than the surrogate CNNs using for manipulation parameterization classifiers.   As a result,  the attacker cannot directly use the  victim CNN or obtain an identical copy to train the attack.  The anti-forensically attacked images produced by the attacker should transfer to fool the investigator's CNN.

To mimic this scenario,  we used \textit{``A-set"} and created two new training data sets to train new surrogates CNNs and the proposed attack.  The new training data sets were created using  four parameters for each manipulation.  The excluded manipulation parameters are  $\sigma=1.5$ for additive Gaussian white noise,  $\sigma=2.5$ for Gaussian blurring,  and $7\times 7$ filter size for median filtering.  The training data created for the proposed attack follows the rules of pixel to pixel correspondence.  

First we evaluated the performance of the newly trained surrogate CNNs on \textit{``eval-set“}.  The testing data were created using four parameters for each manipulation.  The classification accuracies of newly trained surrogate CNNs are shown in Table~\ref{tab: accuracy_3m4p}.  On average,  we achieved the classification accuracy of $98.69\%$ for manipulation detection,  $98.45\%$ for manipulation classification and $83.25\%$ for manipulation parameterization.  These surrogate CNNs were used for building the ensemble to train the proposed attack.  

To attack victim CNNs trained on \textit{``I-set"},  we adopted the same strategies  as in the training data mismatch scenario and built different ensembles of surrogate CNNs for attacking VGG-19 CNNs and other CNNs,  due to the limitation of computer memories.    Then we evaluate the performance of the proposed attack on \textit{``eval-set"}.   The first  testing set  contains  $45,000$ image patches created using  all manipulations and parameters.  The second testing set contains $9, 000$ image patches created  only using manipulation parameters unseen by the attacker.  

The attack success rates of fooling each victim CNN are shown in Table~\ref{tab: para_mismatch}.   For the testing image patches containing all manipulation and parameters,   on average,  we achieved attack success rate of $0.89$ for manipulation detection,  $0.75$ for manipulation classification,  and $0.63$ for manipulation parameterization.  It means that $89\%$ anti-forensically attacked images produces by our attack can fool manipulation detection classifiers,  $75\%$ can fool manipulation classification classifiers,  and $63\%$ can fool manipulation parameterization classifiers.  The mean of attack success rates regardless of class definition was  $0.76$.  It means that $0.76$ anti-forensically attacked images can fool any forensic CNN.  

For the testing image patches containing only attacker's unseen manipulation parameters,  on average,  we achieved the attack success rate of $0.87$ for manipulation detection,   $0.72$ for manipulation classification,  and $0.62$ for manipulation parameterization.   The mean of attack success rates regardless of class definition was $0.74$.  The results indicate once the attack is trained,  the performances on attacking manipulated image patches created with seen and unseen parameters are comparable.  Therefore,  if the attacker knows about the attack success rates of the attack for attacking manipulated images created with known parameters,  the attacker can also use the attack on unknown parameters without performance decrease.

\subhead{The Effect of Missing Training Parameters} Comparing with the attack success rates with  the training data mismatch in Table~\ref{tab: data_mismatch}.   We note that victim CNNs built with TransferNet and PHNet architectures were not influenced much when the attack was trained with less parameters.  It may indicate that these architectures are less sensitive to parameters,  and potentially less robust to the anti-forensic attack.  For other victim CNNs such as the manipulation  parameterization of MISLnet and DenseNet\textunderscore BC,  the attack success scores dropped significantly.  It may indicate that these CNN architectures are very sensitive to the training parameters. 

\subhead{Visual Quality and  Inspection}  We evaluated the image quality of the anti-forensically attacked images produces by our proposed attack  by calculating the mean PSNR and the mean SSIM between manipulated image patches and generated anti-forensically attacked image patches.  The mean PSNR was  $45.83$ and the mean SSIM was $98.91$.  The results  show that  in the training data and parameter mismatch scenario,  the anti-forensically attacked images produced by our proposed attack have  high visual quality.

\begin{figure}[t!]
	\centering
	\includegraphics[scale=0.34]{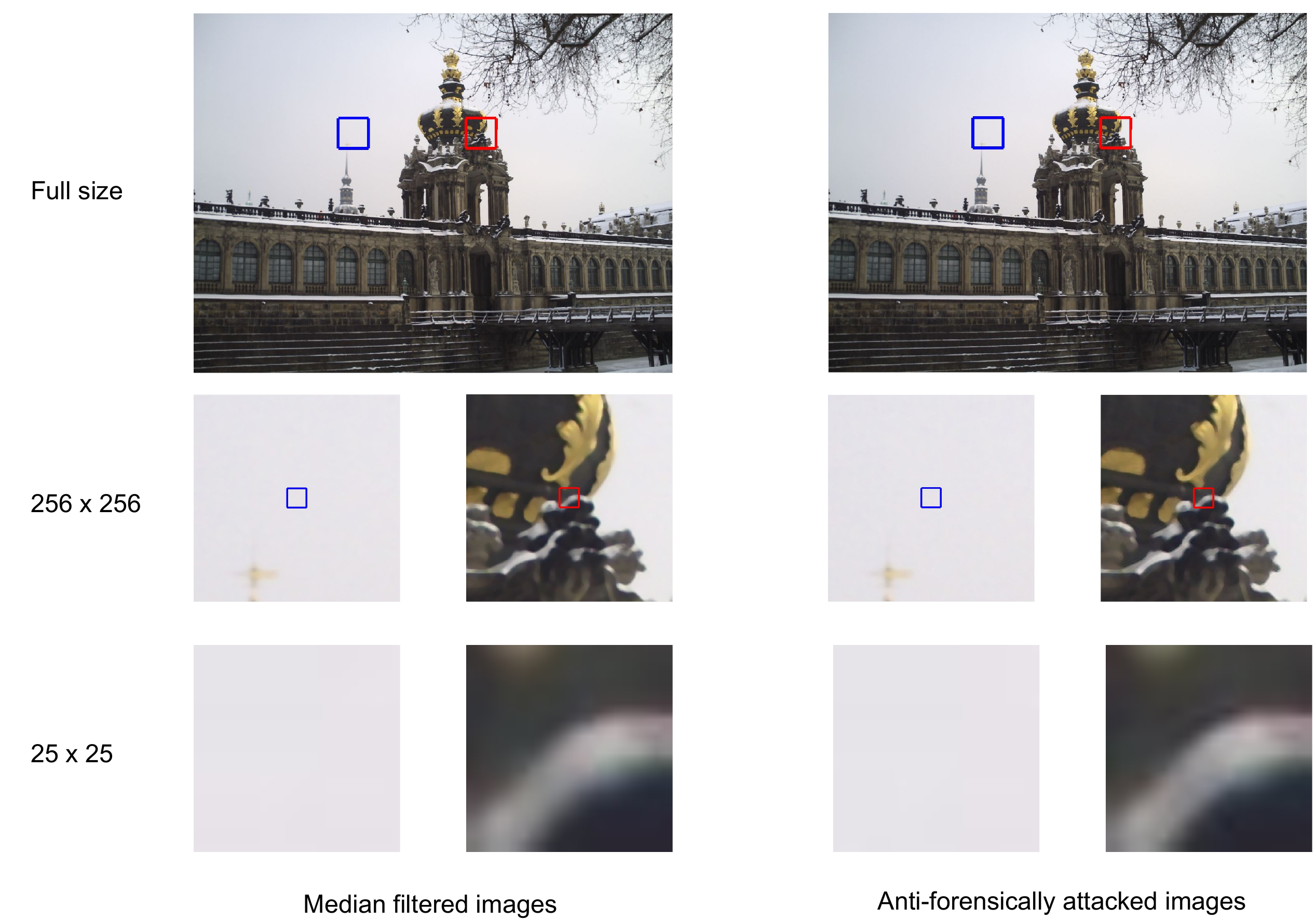}
	\caption{Comparison between median filtered image and its anti-forensically attacked copy produced by our attack.  Left column: an image median filtered using $7\times 7$ filter.  Right column:  anti-forensically attacked images produced by the proposed anti-forensic GAN.   The blue box shows the smooth region of the images,  and the red box shows the textured region of the images.  }
	\label{fig: para_mismatch}
\end{figure}

For visual inspection,  we demonstrated an an-forensically attacked image produced by our proposed attack in Fig. ~\ref{fig: para_mismatch}.   The manipulated image was created using median filtering with $7\times 7$ filter size.  The generator used to create the anti-forensically attacked image was not trained using this parameter.  The blue box shows that smooth region of the images,  and the red box shows the textured regions of the images.  By  inspecting the boxed regions in the zoomed view,  we cannot notice any visual difference  even on the pixel level.   It indicates that it is impossible for the investigator to know whether an image was anti-forensically attacked or not via visual inspection,  if only presented with an image. 

%% file: sections/arch_mismatch_v2_MCS.tex
\vspace{-1em}
\subsection{Training Data and CNN Architecture Mismatch}
\label{subsec: arch_mismatch}
In this experiment,  we evaluated the performance of our proposed anti-forensic GAN attack when the attacker has zero information about investigator's CNN and  has no access to the investigator' s training data.  
In this scenario,  we assume that the attacker knows that the investigator will use a CNN to authenticate image, as well as which manipulations and parameters the CNN will be trained to detect.  However, the investigator will keep their CNN architecture secret and the attacker cannot gather information about it by any means.
As a result,  the attacker  cannot utilize the victim CNN, obtain an identical copy, or probe the output of the victim CNN to train the attack.  Moreover,  the attacker cannot use CNNs built with the same architecture of the investigator's CNN to train the attack.  
It is the attacker's goal for attacked images to have sufficient transferrability to fool the investigator's CNN. 

To experimentally replicate this scenario, 
we treated each CNN trained on the \textit{``I-set"} as a victim CNN to fool.  We used the CNNs trained on the \textit{`` A-set"} as the surrogate CNNs available to the attacker. 
In each case, the ensemble of surrogate CNNs used to train the attack did not contain a surrogate  CNN with the same architecture as the victim CNN.  
 %
%
 Thus we ensured that the attacker did not use investigator's CNN architecture.  Then we used the victim CNN to classify the anti-forensically attacked images produced by the attack.  
%
As before, the size of VGG-19 required us to slightly modified our training ensemble due to GPU memory limitations. 
To attack VGG-19 CNNs,  the training ensemble  was built using  surrogate CNNs of all other architectures.   To attack other victim CNNs,  
the ensemble was built using surrogate CNNs of other architectures excluding architectures of VGG-19 and the victim CNN.  The generated anti-forensically attacked images then were classified by  each victim CNN. 

We evaluated the performance of our proposed anti-forensic attack on $45,000$ manipulated image patches from the \textit{``eval-set''}.  The attack success rates for fooling each victim CNN are shown in Table~\ref{tab: arch_mismatch}.  On average,  our proposed anti-forensic GAN attack achieved the attack success rate of $0.89$  for manipulation detection,  $0.71$  for manipulation classifcation,  and $0.68$ for manipulation parameterization.  The mean of attack success rates regardless of the class definition was  $0.76$. 

These results demonstrate that even when the attacker has zero information about the victim CNN, including its architecture,  
our proposed anti-forensic GAN attack can still fool the investigator's CNN.  
This is important because it shows that our attack can be launched in a realistic black box scenario.  
Moreover,  our finding demonstrates that  transferable attacks against forensic CNNs can be achieved,  unlike what researchers have previously found~\cite{transferability_Barni, zhao2020effect}.  While this attack is not perfect,  the successful attack rates are high enough to pose a real threat to CNN-based forensic algorithms.


\begin{table}[t!]
	\centering
	\caption{Attack succes rates (ASRs) achieved by the proposed anti-forensic GAN attack in the training data and CNN architecture mismatch scenerio. }
	\resizebox{0.50\textwidth}{!}{
	\begin{tabular} {|l cccc|}
		\hline
		 \multicolumn{5}{|c|}{\textbf{Proposed Anti-Forensic GAN Attack}}\\		
		\hline

 \textbf{CNN Architect. } & \textbf{Detection} & \textbf{Classification}  &\textbf{ Parameterization} &\textbf{Avg.}\\
		MISLnet& 0.99&0.72&0.95&0.88\\
		TransferNet& 0.99 & 0.80 & 0.91 &0.90\\
		PHNet & 0.64& 0.98 &0.78&0.80\\
		SRNet& 0.85& 0.03 & 0.45&0.44\\
		DenseNet\textunderscore BC & 0.94& 0.77&0.09 &0.60\\
		VGG-19& 0.94&0.98 &0.92&0.95\\
		\textcolor{blue}{\textbf{Avg.}} & \textcolor{blue}{\textbf{0.89}}& \textcolor{blue}{\textbf{0.71}} &\textcolor{blue}{ \textbf{0.68}}&\textcolor{blue}{\textbf{0.76}}\\\hline
	\end{tabular}}
\label{tab: arch_mismatch}
\end{table}

\subhead{Visual Quality} We evaluated the image quality of the anti-forensically attacked images produces by our proposed attack  by calculating the mean PSNR and the mean SSIM between manipulated image patches and generated anti-forensically attacked image patches.  The mean PSNR $46.76$ and the mean SSIM$46.76$.  The results show that in the training data and CNN architecture mismatch scenario,  our proposed  attack  will not introduce perceptible distortion to the anti-forensically attacked image.


%% file: sections/conclusion.tex
\label{sec: conclusion}
In this paper, we proposed a new anti-forensic  GAN attack to fool CNN-based forensic algorithms.   Different from standard GANs,  our proposed anti-forensic GAN attack uses an ensemble of surrogate CNNs to enforce the generator to learn comprehensive aspects of forensic information about real unaltered images,  and produces anti-forensically attacked images that can fool forensic CNNs.  Furthermore,   our proposed attack demonstrated strong transferability to fool CNNs not used during training under various limited knowledge scenarios.  We conducted a series of experiments to evaluate our proposed anti-forensic GAN attack.  We showed that the proposed attack can fool many state-of-art CNN-based forensic algorithms in the perfect knowledge scenarios  and other scenarios when the attacker lacks the information of the  investigator's CNN and training data.  Additionally,  we showed that under all scenarios our proposed attack leaves behind no visual distortions to the produced anti-forensically attacked image.   Hence the investigator cannot uncover the forgery by mere visual inspection.